\documentclass[sn-mathphys-num]{sn-jnl}

\usepackage{graphicx}
\usepackage{multirow}
\usepackage{amsmath,amssymb,amsfonts}
\usepackage{amsthm}
\usepackage{mathrsfs}
\usepackage[title]{appendix}
\usepackage{xcolor}
\usepackage{textcomp}
\usepackage{manyfoot}
\usepackage{booktabs}
\usepackage{algorithm}
\usepackage{algorithmicx}
\usepackage{algpseudocode}
\usepackage{listings}

\usepackage[most]{tcolorbox}

\usepackage{color}
\usepackage{soul}
\definecolor{lightyellow}{rgb}{0.97, 0.97, 0}
\sethlcolor{lightyellow}

\usepackage{placeins}

\begin{document}

	\title{The Need for Guardrails with Large Language Models in Medical Safety-Critical Settings: An Artificial Intelligence Application in the Pharmacovigilance Ecosystem}
	
	\author[1]{\fnm{Joe B.} \sur{Hakim}}\email{joehakim@mit.edu}	
	\author[2]{\fnm{Jeffery L.} \sur{Painter}}\email{jeffery.l.painter@gsk.com}
	\author[3]{\fnm{Darmendra} \sur{Ramcharran}}\email{darmendra.x.ramcharran@gsk.com}
	\author[6]{\fnm{Vijay} \sur{Kara}}\email{vijay.x.kara@gsk.com}
	\author[2]{\fnm{Greg} \sur{Powell}}\email{gregory.e.powell@gsk.com}
	\author[4]{\fnm{Paulina} \sur{Sobczak}}\email{paulina.a.sobczak@gsk.com}
	\author[5]{\fnm{Chiho} \sur{Sato}}\email{chiho.2.sato@gsk.com}
	\author[6,7]{\fnm{Andrew} \sur{Bate}}\email{andrew.x.bate@gsk.com}
	\author*[8]{\fnm{Andrew} \sur{Beam}}\email{Andrew\_Beam@hms.harvard.edu}

	\affil*[1]{\orgname{Harvard-MIT Department of Health Sciences and Technology}, \orgaddress{\city{Cambridge}, \state{MA}, \country{USA}}}	
	\affil[2]{\orgname{GlaxoSmithKline}, \orgaddress{\city{Durham}, \state{NC}, \country{USA}}}
	\affil[3]{\orgname{GlaxoSmithKline}, \orgaddress{\city{Providence}, \state{RI}, \country{USA}}}	
	\affil[4]{\orgname{GlaxoSmithKline}, \orgaddress{\city{Warsaw}, \country{Poland}}}
	\affil[5]{\orgname{GlaxoSmithKline}, \orgaddress{\city{Tokyo}, \country{Japan}}}	
	\affil[6]{\orgname{GlaxoSmithKline}, \orgaddress{\city{Brentford}, \state{Middelsex}, \country{UK}}}
	\affil[7]{\orgname{London School of Hygiene and Tropical Medicine}, \orgaddress{\city{London}, \country{UK}}}
	\affil*[8]{\orgname{Department of Epidemiology, Harvard T.H. Chan School of Public Health}, \orgaddress{\city{Boston}, \state{MA}, \country{USA}}}

	
	\abstract{Large language models (LLMs) are useful tools with the capacity for performing specific types of knowledge work at an effective scale. However, LLM deployments in high-risk and safety-critical domains pose unique challenges, notably the issue of ``hallucination,'' where LLMs can generate fabricated information. This is particularly concerning in settings such as drug safety, where inaccuracies could lead to patient harm. To mitigate these risks, we have developed and demonstrated a proof of concept suite of guardrails specifically designed to mitigate certain types of hallucinations and errors for drug safety, and potentially applicable to other medical safety-critical contexts. These guardrails include mechanisms to detect anomalous documents to prevent the ingestion of inappropriate data, identify incorrect drug names or adverse event terms, and convey uncertainty in generated content. We integrated these guardrails with an LLM fine-tuned for a text-to-text task, which involves converting both structured and unstructured data within adverse event reports into natural language. This method was applied to translate individual case safety reports, demonstrating effective application in a pharmacovigilance processing task. Our guardrail framework offers a set of tools with broad applicability across various domains, ensuring LLMs can be safely used in high-risk situations by eliminating the occurrence of key errors, including the generation of incorrect pharmacovigilance-related terms, thus adhering to stringent regulatory and quality standards in medical safety-critical environments.}

	
	\maketitle

	\section{Introduction}

	The integration of large language models (LLMs) into the fabric of numerous applications has positioned them as instrumental in navigating the complex challenges in biology and medicine \cite{tang2023evaluating}. The breadth of their application, combined with their rapid evolution, has created anticipation that LLMs will be near-universal solvers across the biomedical landscape \cite{tang2023evaluating, singhal2022large, clusmann2023future}. Yet, alongside this growing optimism, there is an increasing cognizance of their limitations that may impede their applicability in specific areas of scientific inquiry. Prominently, the phenomenon of ``hallucinations'' -- instances of generating baseless information -- stands as a pivotal concern \cite{zhang2023siren}. This phenomenon is a byproduct of the mechanisms underpinning LLMs, which rely implicitly on internally stored ``memories'' for response generation, without explicit grounding in verifiable facts \cite{mckenna2023sources}. LLMs also face challenges in communicating the uncertainties of their outputs to end-users effectively. Though measures of uncertainty can sometimes be quantified, validating the trustworthiness of LLM outputs remains a challenge \cite{xiong2023can, wagle2023empirical} including within the biomedical domain \cite{bolton2024rambla}.

	In contexts where inaccuracies can result in severe consequences, particularly in decision-making processes affecting patient safety, the issue of LLM hallucinations and omission of key information \cite{ema2018} becomes acutely significant \cite{bowen1993safety}. One critical domain is drug safety, also known as pharmacovigilance (PV), which involves the ongoing surveillance for adverse events (AEs) linked to pharmaceutical medicines and vaccines\cite{world2002importance}. Given the limitations of pre-market trials in fully characterizing a drug or vaccine's safety profile, PV relies on the collection and analysis of spontaneously reported AEs, vital for continued assessment of a product's benefit-risk. The reported information is transcribed into an Individual Case Safety Report (ICSR) which serves as the standardized international framework for AE reporting, encompassing a vast array of information sourced globally in varied formats and demanding timely review and processing. The non-random process by which ICSRs are collected, coupled with the prevalence of incomplete or erroneous data, underscores the necessity for clinical review to unearth potential safety signals for further exploration and serve as the primary data source to formally evaluate a potential causal association \cite{bate2009quantitative}. Consequently, a challenge within PV lies in the efficient parsing of extensive, noisy, and often incomplete domain-specific textual data, some of which may be contradictory, to identify safety signals meriting additional investigation.

	The propensity of LLMs to hallucinate and omit key details presents a considerable hazard if applied naively within the PV domain, which is inherently safety-critical. For instance, an LLM might erroneously suggest that an ICSR details a serious AE such as liver failure while this is not mentioned in the source report, potentially signaling a false-positive safety concern and diverting resources from legitimate safety investigations. Moreover, understanding how LLMs are integrated with human end-users becomes essential, as human-mediated oversight systems will likely remain indispensable for certain tasks within safety-critical applications for the foreseeable future.

	Preventing and mitigating hallucinations involves the implementation of ``guardrails'' around LLMs to shape and restrict their output. While the term guardrail lacks a precise definition in this context, it is here understood as a series of constraints applied to either an aspect of the LLM or its output to ensure adherence to predefined criteria. One approach involves ``structural guardrails,'' defined as mechanisms ensuring model outputs maintain a consistent structure (e.g., CSV, XML, JSON) \cite{dong2024building}, thus obviating the need for further processing of free text to extract pertinent information.

	This paper focuses on ``semantic guardrails,'' aimed at verifying the accuracy of LLM output by checking for biased or problematic content and coding errors. These guardrails may be ``hard,'' offering clear binary outcomes, or ``soft,'' providing probabilistic assessments regarding the potential error in the output\footnote{\url{https://github.com/guardrails-ai/guardrails}}. Within pharmacovigilance, such guardrails are pivotal in enforcing the avoidance of errors that may impact safety decisions resulting in patient harm analogous to medical ``never events'' incidents in clinical practice contexts identified by U.S. and U.K medical organizations as wholly preventable and unacceptable \cite{hsjNeverEvents}. These never events, deemed intolerable and preventable, have the potential to lead to significant harm or mortality and usually trigger comprehensive investigations to avert recurrence. Examples include severe allergic reactions to contraindicated medications or dosing errors and are ``serious incidents that, due to the provision of systemic protective barriers at a national level, are completely preventable and should have been preemptively addressed by all healthcare providers'' \cite{anderson2020using}, analogous to guardrails. Hence, to function within safety-critical domains like PV, semantic guardrails must ensure the absolute prevention of defined ``never event'' errors that have the potential to adversely impact pharmacovigilance decision-making \cite{nqfSres}.

	In our investigation, we introduce a comprehensive set of both hard and soft semantic guardrails designed to enable LLMs to function within the high-risk, safety-critical environment of PV. Focusing on the complex and expansive data processes integral to PV, our research specifically addresses the challenge of processing multilingual ICSR intake and analogous processing within a real-world PV system. This encompasses a text-to-text task that involves both structured to unstructured data conversion and translation. Our guardrails were specifically tested on the task of transforming Japanese language ICSRs (combined unstructured and structured, tabular data with numeric codes for various biomedical concepts) into English narrative text for subsequent analysis by safety professionals.

\begin{figure}
			\centering
			\includegraphics[width=5.0in]{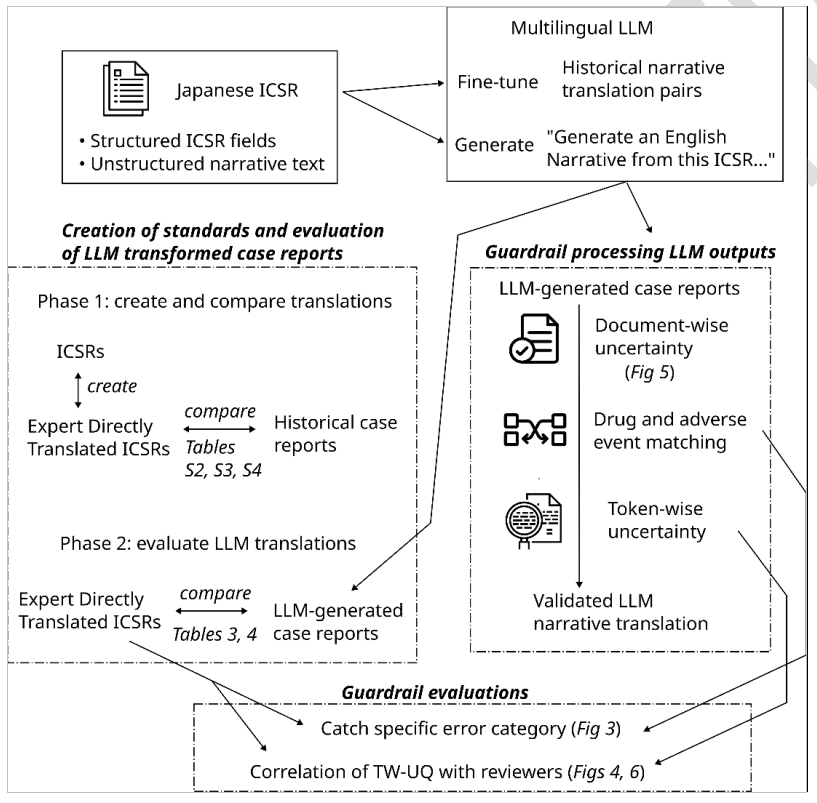}
			\caption{Graphical summary of the large language model (LLM) workflow. We used extra structured fields and unstructured narrative texts from individual case safety reports (ICSRs), along with historical matched language examples, to fine-tune an LLM. We added a specific task prefix, and generated an English narrative from a Japanese ICSR, and finally checked this process via several guardrails: the document-level uncertainty, drug and adverse event matching, and token-level uncertainty guardrails (see Methods section).}\label{fig_01}
		\end{figure}

	We identified multiple potential failure modes for LLMs within this context and engineered a series of guardrails to mitigate these risks (Figure \ref{fig_01}). We implemented a hard semantic guardrail to address model outputs with generated drug or vaccine names not present in the source text, utilizing existing drug safety dictionaries and tools to ensure consistency of key drug- and vaccine-related information between the source text and the LLM-generated English narrative. Additionally, we incorporated two soft semantic guardrails to communicate the model's uncertainty regarding the quality and accuracy of both the input text and its final translation, thereby flagging instances potentially requiring further human review. While our study concentrates on a critical, real-world case in PV, we posit that the framework developed herein holds relevance across a multitude of medical safety-critical domains.
	
	\section{Methods}
	
		A schematic of the workflow is presented in Figure \ref{fig_01}, including processing the ICSRs, the LLM tasks, creation of standards and the evaluation of LLM generated case reports, the sequential guardrail processing, and the evaluations of the guardrails.
			
		\subsection{Data Acquisition}

			The dataset utilized in this study was sourced from GSK's global safety database as part of a collaboration by providing Harvard University, Cambridge, Massachusetts, USA, access on a privately maintained, secure server equipped with advanced graphics processing units (two 80 GB A100s). This dataset encompasses over two decades of ICSRs, with more than 4 million cases available for review. For the purposes of our assessment, the analysis concentrated on the original ICSRs as submitted to GSK, prior to any form of human review. This excluded any subsequent modifications or additional data reported post-initial submission, including follow-up details.
			
		\subsubsection{Analysis of Individual Case Safety Reports}

			Spontaneously reported AEs are transcribed into an Individual Case Safety Report (ICSR) which serves as the standardized international framework for AE reporting.  A valid ICSR for entry into the GSK Global safety database is comprised of four essential elements: (1) at least one identifiable reporter; (2) an identifiable patient; (3) at least one suspect adverse reaction; and (4) at least one GSK suspected product \cite{ema2017}. Pharmaceutical entities often accumulate reports in large volumes from various data partners. Whenever feasible, these reports are exchanged using standardized E2B XML documents \cite{fdae2b}, which offer structured fields alongside narrative descriptions of each case. For our study, we treated the entirety of information initially available as a singular data point. This approach included aggregating additional structured fields such as the country of the primary source, country of occurrence, level of seriousness (including death, life-threatening situations, hospitalization, disability, congenital anomalies), relevant dates, details of the reporter (name, organization, country), patient demographics (age, sex), primary reaction of the patient, and the implicated medicinal product.
			
		\subsection{Development of a Multilingual Corpus for LLM Pretraining}

			We constructed a multilingual corpus of ICSRs to serve as the dataset for text-to-text fine-tuning of an LLM. To achieve this, we aligned the raw text from the submitted ICSRs (source text) with the human-generated summaries provided by a third-party contractor (original standard target text) to create text pairs in four languages: Japanese, Spanish, French, and German. These languages were selected due to their prevalence in the database and, particularly for Japanese, the complexity they present in translation tasks. While our analysis primarily concentrates on Japanese due to the high number of ICSRs available in this language, the LLMs are designed for multilingual application. To integrate the additional structured fields, we prefixed each one to the source text of the ICSR in the format:
			
			\begin{verbatim}
	field_name_1: field_value_1; field_name_2: field_value_2
			\end{verbatim}

			To fine-tune an LLM or employing it in text generation, we also prefixed a brief instruction indicating the specific task for the model, such as ``Translate the following Japanese case report into English narrative text'' for translations from Japanese to English. Our pretraining corpus was enriched further with direct translation pairs from the OPUS-100 corpus \cite{zhang2020improving}, a comprehensive multilingual translation dataset covering 100 languages, thus furnishing additional examples for model fine-tuning on translation tasks involving parallel language sets. The volume of pretraining examples is detailed in Table \ref{t1}.
		
		\subsection{Development of the ICSR translation LLM}
		
			\subsubsection{Model fine-tuning and generation}
		
				
				In our study, we conducted an evaluation of three LLMs with parameter sizes ranging from 700 million to 7 billion: \mbox{mt5-xl}\footnote{\url{https://aclanthology.org/2021.naacl-main.41.pdf}}, \mbox{mpt-7b-instruct}\footnote{\url{https://huggingface.co/docs/transformers/main/en/model_doc/mpt}}, and \mbox{stablelm-japanese}\footnote{\url{https://huggingface.co/stabilityai/japanese-stablelm-base-alpha-7b}}.  The criteria for selecting these models included the relevance of their initial pretraining objectives, the scale of the models, and the computational resources required for their operation. These models underwent further fine-tuning for translation tasks, utilizing a corpus composed of 131,037 examples from ICSRs and texts from the OPUS-100 dataset (Table \ref{t1}). The training process was applied uniformly across Japanese, Spanish, French, and German, adopting a split of 70\% for training, 15\% for validation, and 15\% for testing. This distribution ensured a balanced representation of languages and sources (ICSR vs. OPUS-100) within each set.
				
				For the generation phase, we evaluated a variety hyperparameters. Utilizing beam search \cite{graves2012sequence}, we experimented with different settings for the temperature (0.5, 0.7, 1.0, 2.0) and beam counts (3, 10, 25). Additionally, in our application of contrastive search \cite{su2022contrastive}, adjustments were made to the $\alpha$ values (0.2, 0.6, 0.9) and the top-k selections (4, 8, 16, 64). The optimal set of generation hyperparameters was determined based on the BLEU score \cite{papineni2002bleu} performance on the validation set, ultimately selecting a contrastive search configuration with $\alpha = 0.2$ and top-k = 16.

			\subsubsection{Model evaluation}

				In our initial assessments, we concentrated on evaluating the Japanese translation quality, a task of significant relevance in PV due to the human resources required with securing proficient translators for Japanese drug safety data. We conducted comparative analyses of the three models, utilizing per-token perplexity as a metric on a validation subset comprising 7,820 ICSRs, which constitute approximately 13\% of the total Japanese ICSRs in our dataset. For the best performing model, we further explored its translation capabilities by applying standard machine translation evaluation metrics, including the BLEU score \cite{papineni2002bleu}, SACRE-BLEU score \cite{post2018call}, and word error rate \cite{klakow2002testing}.

			\subsection{Expert human evaluation of the target text}

				After finalizing our model, we performed a comprehensive evaluation aimed at assessing its efficacy in translating cases that were originally documented in Japanese. This analysis involved 210 cases, all sourced from Japan and initially documented in Japanese. The selection of these cases was governed by a predefined set of criteria. Our goal was to achieve an even distribution across various product categories, with our sample evenly divided among vaccines, general medicines, and specialized products, like those in oncology. Priority was given to serious cases that had been subjected to in-depth analysis upon their reception, thus offering a comprehensive insight into potentially critical incidents. Additionally, we sought to maintain a balanced representation of products across these categories. The cases spanned the entire 20-year period for which we had data, ensuring temporal representativeness. Finally, our case selection employed random sampling within these specific strata to reflect the overall distribution of Clinical Utility Score for Prioritization (CUSP) scores \cite{kara2023finding} found in our entire ICSR database. This methodology was designed to secure a broad and diverse representation in the completeness of the cases under review.

			\subsection{Phase 1: Establishment of High-Quality Baseline Translations}

				The first phase was dedicated to creating a baseline foundation of high-quality translations. Each of the 210 Japanese ICSRs, available in the database as previous translations into English by an external contractor, was subjected to a thorough review by two independent PV experts fluent in both Japanese and English. This double-blind review not only verified the translations for accuracy and fluency, but also established a robust English ``ground truth'' for further comparative analysis. The outcomes from this phase's evaluation are detailed in the supplementary materials, with Tables \ref{tab:s2}, \ref{tab:s3}, and \ref{tab:s4} offering a juxtaposition of the initial standard target texts against the evaluations conducted by the bilingual PV specialists.
				
			\subsection{Phase 2: Evaluation of LLM Translations Against Established Baseline}

				In the next phase, we assessed the LLM-generated English translations against the ``ground truth'' translations derived in Phase 1 of the experiment. This assessment was carried out by PV experts proficient in English, with experience in safety evaluations. Employing a carefully designed evaluation framework, they conducted independent dual reviews of each translation, incorporating both a detailed five-category assessment system (Table \ref{tab:s1} for category specifics) and binary evaluation criteria (Table \ref{tab:s4}). In instances of binary evaluation, the presence of any noted error category, observed even once, warranted its marking, with evaluators having the option to detail the specific nature of the error. Moreover, the experts assessed the clinical acceptability of each processed ICSR for reporting to regulatory agencies. Any discordance among the evaluations was resolved by an additional independent senior expert. For the four-category criteria, evaluations were made on a five-point Likert scale \cite{likert1932technique}, with ratings ranging from 1 (least favorable) to 5 (most favorable), as detailed in Table \ref{tab:s1} in the supplement for the definitions of each rating level.

				To streamline the evaluation, a custom web application was created, affording the reviewers the ability to methodically compare translations side-by-side and to log their assessments using dropdown menus and open-ended text fields. Cases were randomly distributed among a team of reviewers to minimize the potential for individual reviewer and selection bias. This application was designed with tracking capabilities for capturing individual evaluator responses, and it was programmed to automatically signal for independent expert adjudication should discrepancies between reviewers emerge.

		\subsection{LLM guardrails for ICSR translations}
		
		We developed one hard and two soft semantic guardrails for this application, as described below in order of application in the ICSR processing pipeline:
				
		\subsubsection{Document-wise uncertainty quantification (DL-UQ)}

			This soft guardrail identifies submitted documents that are unlikely to be ICSRs reports (based on statistical probabilities as reported by a model, as opposed to using the 4 aforementioned validation criteria for ICSRs). To support potential automation of ICSR intake, this guardrail detects documents unlikely to be an AE report and prevents any LLM processing of these reports. The DL-UQ guardrail first creates a document level embedding by performing an average pooling operator to the token-level embeddings created using the source language encoder LLM. Next, a k-nearest neighbors' Euclidean distance is calculated between the embedding for the submitted document and a cache of ICSR embeddings created using the same methodology from the training data. This distance is a measure of uncertainty according to the LLM as it measures how anomalous a new submission is relative to the documents the model has seen before and can be used to automatically discard a submission or flag it for review. A distance threshold can be tuned to achieve a desired trade-off between sensitivity and specificity.
		
		\subsubsection{MISMATCH (drug and AE mismatching)}

			This hard guardrail enforces a ``never'' event by identifying drug names that appear in either the source text or target text but not both, indicating that a drug name has been either mistranslated or hallucinated. This kind of error represents a so-called ``never event'' because incorrectly identifying a drug in an ICSR could have dire safety consequences and should be avoidable. To implement this guardrail, we matched (with regular expressions) both the source and target texts for any mentions of drugs; similarly, this was implemented for AEs. Then, we used two dictionaries (a custom in-house drug dictionary from the global safety database, and MedDRA, Medical Dictionary for Regulatory Activities \cite{brown1999medical}; 28K preferred terms) to find the matching terms, and whether the set difference had any elements corresponding to unmatched terms. The dictionary matches allowed generic-trade name associations for drugs. Note: this guardrail did not match terms that are slightly misspelled drug names or AEs, since those are not matched by the regular expression-based text matching comparison with terms in the dictionaries. If there was a mismatch, this hard guardrail would trip and the eventual integrated system would route outside of the standard case processing and for further adjudication, either through post-processing or human-in-the-loop assessment and correction.

		\subsubsection{Token-wise uncertainty quantification (TL-UQ)}
		
			This soft guardrail identifies potential LLM errors at word and sub-word levels. Each token in the vocabulary is assigned a log probability by the LLM, and we take the entropy of this multinomial distribution as the token-level uncertainty score. Intuitively, the more entropy in the predictive distribution of the next token, the ``less certain'' the model is in generating that specific token.		
		
		\subsection{Guardrail assessments}

			We assessed each guardrail as follows:
			
			For DL-UQ, using the train validation split described above (see ``Data pre-processing'' and ``Model evaluation''), we sampled 80 example texts from the training and validation sets, and produced a score for each. We then injected a sample of 25 ``extraneous samples,'' which included 14 Japanese Wikipedia articles, 7 Japanese fake case reports (in a similar format as the original case reports), 2 Japanese texts that have nothing to do with PV, and 2 non-Japanese texts. We plotted the numeric score for each example to evaluate the separation and reported the area under the receiver operator curve (AUROC) for a discrimination between validation and extraneous samples.

			For MISMATCH, the primary evaluation of this guardrail was whether the specific targeted ``never event'' is always flagged when the target text contains that error. To this end, we used the human evaluators' flagged drug errors as the exemplar never events on a (programmatically) randomly selected sample of 20 cases. We calculated the fraction of cases caught by the MISMATCH guardrail where the human evaluators indicated a drug name had been hallucinated spontaneously. The MISMATCH guardrail was also useful for the other categories. We divided its fixes in the ``generic-trade name'' category by how the specific drugs mentioned were flagged by the MISMATCH guardrail itself (``fixed by mismatch guardrails'') or by a separate system that we added to check if generic-trade names match by looking for parentheses (``direct generic-trade name checking''). For this, we used the existing pairs of generic-trade names from the dictionary afterwards. We divided fixes in the ``drug spelling issues'' section by whether they were directly fixed by the guardrails (``fixed by mismatch guardrails'') or not fixed, which happened when the same drug had both correct and incorrect spellings in the LLM generated output (``multiple mentions''). In these cases, the guardrail did not find the misspelled drug by matching the text in the LLM output in English, and it did not detect that the drug mention in the Japanese source text is unmatched, because the drug is also spelled correctly. The frequency of the MISMATCH guardrail flagging individual drug and AE names is evaluated by a ``missrate,'' which is a measure of the frequency of these erroneous outputs that are not fixed by this guardrail. A missrate of 1.0 indicates that, for example, the source text contained a number of drugs or AEs that are not matched to any translated terms in the target text. In the standard translations used to train the model, we expect this to be 0.0, so any missrates $>$ 0.0 are due to a limitation of the drug or AE lists, or a misspelling of these terms.

			For evaluation of the TL-UQ guardrail, we showed a qualitative example of a visualization flagging spans of uncertain text. In that example, we correlated the flagged spans with a human evaluator's assessment of specific errors in that case. Spans were flagged by differing intensities of text highlighting, from least to most, corresponding to the 10th percentile, 5th percentile, and 1st percentile most entropic predicted tokens. Quantitative evaluations were conducted by stratifying each reviewed case by ``Is the case clinically accurate'', ``Wrong name or information'', and ``Incorrect AE/Wrong outcome'' and assessing the case entropy score (an average of the individual token entropies) for each case in each category.
		
	\section{Results}

		\subsection{Translation model development and evaluation}
	
			We considered three LLMs, that at the time of beginning this study, were performant multilingual models that could run given local resources on our internal servers: mt5-xl, MPT-7B, and stablm-japanese. We first assessed how well each could translate ICSRs from Japanese to English without any task-specific fine-tuning and then assessed this ability when the models were fine-tuned with ICSR data (Table \ref{t2}).
	
			These results indicate that none of the base models are suitable for translation ``off the shelf'' (Table \ref{t2}). Fine-tuning improved all models by a significant margin and only mt5-xl reached a suitable perplexity after fine-tuning (Table \ref{t2}). This is most likely due to this base model being pretrained explicitly on Japanese text, while the others likely only encountered Japanese text during their initial pretraining in an extremely small number of instances. On this basis, we decided to move forward with the mt5-xl model for further evaluation.
	
			Traditional metrics of machine translation quality for mt5-xl show a BLEU score of 0.39, which is considered to be associated with relatively high-quality translations \cite{googleEvalModels}, as are the Sacre-BLEU score of 0.44 and the word error rate of 0.73.
				
		\subsection{Phase 1 evaluations: evaluation of existing standard produced target text}
	
			Using the rubric in Supplemental Table \ref{tab:s1}, the reviewers evaluated the quality of the original standard supplied target texts. In Table \ref{tab:s2}, we report summary statistics evaluating whether the standard supplied target text sufficiently captured the same meaning as the source texts. Supplemental Table S3 reports the human-assessed clarity of the source texts and incorporates a two-reviewer system to get an inter-rater agreement in this metric.

			Both rater 1 and rater 2’s median scores were 4.0 (mostly clear and easy to read). Calculating the inter-rater agreement using Cohen’s Kappa, and quadratic weights, gave a Kappa of 0.542. The interpretation of this is typically domain-specific and variable with the number of categories, but in this case represents a much better than random association between the raters and shows consistency in the clarity of the source material.

			Supplemental Table \ref{tab:s4} reports the Phase 1 reviewers’ assessment of the translation accuracy between the standard provided source text and the target text. The human evaluations from the Phase 1 component, in which we checked the set of original data fed into the model (the source ICSR plus the ``ground truth'' translation), show errors and other issues with the input data. The columns represent, e.g. for ``Added information'', that there was additional text in the standard provided source text relative to the target text.

		\subsection{Phase 2 evaluation: expert assessment of LLM produced translations}
	
			We evaluate the LLM produced translations via the human reviewer’s Likert-like criteria (Table \ref{t3}) and binary criteria (Table \ref{t4}). When compared to the existing human translation on the same source text in the database, slightly fewer cases had ``perfect'' (5) clarity scores when generated by LLM (45\% vs 56\% with human translation). For most categories, the translations were rated as 3 or higher, indicating that they were generally considered acceptable. The notable exception concerned correctness of the LLM translation, which was rated 2 for 12.6\%, indicating significant errors that would affect the interpretation (Table 3).

			Following the global assessment of the suitability of the translations, a fine-grained assessment was performed to detect the presence of different error categories in the target text. The results (Table \ref{t4}) showed the LLM translation had errors in categories including dates/times (60\% error rate), drug names (59\% error rate), AEs (66\% error rate), and 62\% had nonsensical phrases, including grammatical errors. In the ``Other errors'' category, the most frequent were typos in drug names, missing causality information, repeated information, incorrect specification of concomitant medications, incorrect inferred indications, incorrect or missing batch number, and other errors that overlapped with those in other categories (e.g. wrong date,).

		\subsection{Assessment of DL-UQ guardrail}
		
\begin{figure}
			\centering
			\includegraphics[width=5.0in]{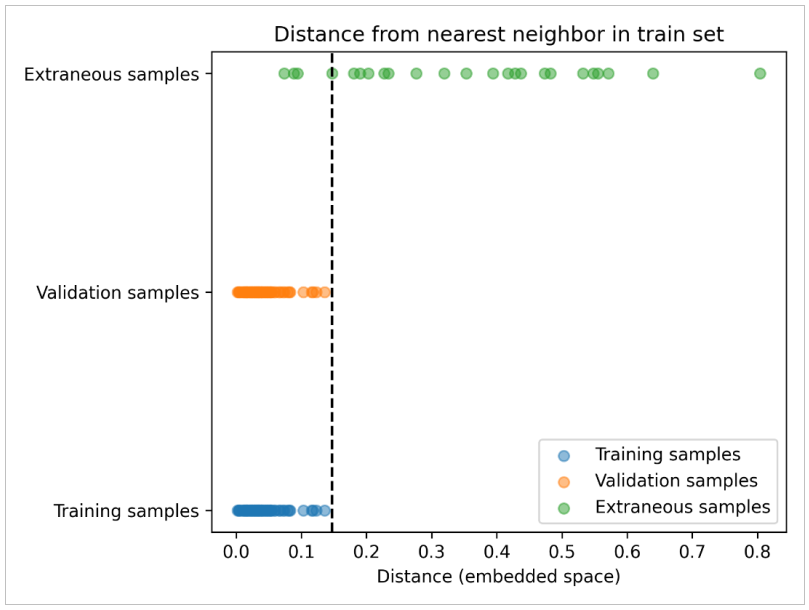}
			\caption{The distribution of document-level uncertainty scores in extraneous, validation, and training samples. The vertical bar represents the minimum validation sample score that is greater than all the validation and training samples. }\label{fig_02}
		\end{figure}

			The DL-UQ metric was applied to training, validation, and non-ICSR Japanese documents. Figure \ref{fig_02} shows that the non-ICSR documents typically had higher distances to their closest training sample in embedding space and, with three counterexamples, can be discriminated from training and validation examples without training.

			The distribution demonstrates separation of the assigned scores for the different categories of cases. Although not completely separated, the separation of the validation and extraneous samples corresponds to an AUROC in the validation data of 0.80.

		\subsection{Assessment of TL-UQ guardrail} 
\begin{figure}
			\centering
			\includegraphics[width=5.0in]{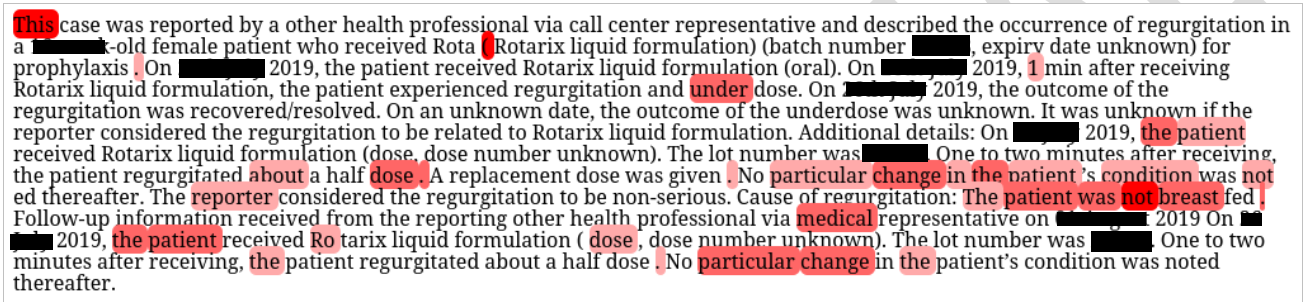}
			\caption{Example flagged spans using the TL-UQ guardrail. Differing levels of red highlighting correspond to increasing relative scores: least color saturation: between 10\textsuperscript{th} percentile and 5\textsuperscript{th} percentile scores for the whole text. Medium color saturation: between 5\textsuperscript{th} and 1\textsuperscript{st} percentile scores. Least color saturation: 1\textsuperscript{st} percentile and above scores. Sensitive information has been redacted with black bars.}\label{fig_05}
		\end{figure}

\begin{figure}
			\centering
			\includegraphics[width=5.0in]{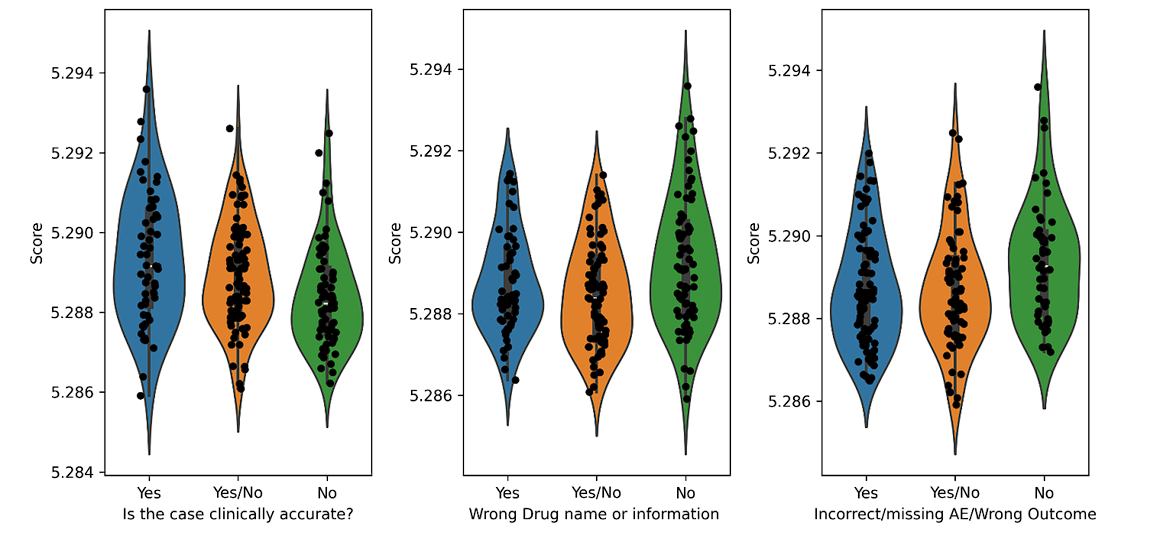}
			\caption{TL-UQ distributions. Stratifying each reviewed case by ``Is the case clinically accurate'', ``Wrong name or information'', and ``Incorrect AE/Wrong outcome'' and reporting entropy score distributions.}\label{fig_06}
		\end{figure}

			An example of a visualization of TL-UQ is shown in Figure \ref{fig_05} and highlights the distribution of the entropy score, which may facilitate efficient and targeted human-in-the-loop review. Figure \ref{fig_06} shows the distributions of TL-UQ uncertainty scores, stratified by clinical accuracy, wrong drug or information, and incorrect/missing AE/wrong outcome. Mann-Whitney U tests (using a Bonferroni correction with n=9 trials) revealed significant differences in in the ``clinical accuracy'' stratification (Yes vs. No, adj. p-value of 0.0031, Yes/No vs. No, adj. p-value of 0.043) and the ``wrong drug'' stratification (Yes/No vs. No, adj. p-value of 0.028). In each of these cases, the trend was the more ``correct'' direction. More clinically accurate, less incorrect drugs/AEs trended towards a higher uncertainty score, implying that entropy correlates negatively with the model’s human evaluated performance. The observed trend of higher entropy scores correlating with more clinically accurate outputs and fewer incorrect drug mentions may be interpreted as counterintuitive. However, one possible explanation is that the distribution of entropy scores reflects inappropriate model confidence, where the model is more confident in its predictions for cases it is more likely to get wrong. Further investigation is needed to fully understand this pattern, but the results suggest that entropy scores, even at the token level, can provide a useful, if subtle, signal of the model's likely correctness on a given case (see Figure \ref{fig_02} for example). 

		\subsection{Assessment of MISMATCH guardrail}

\begin{figure}
			\centering
			\includegraphics[width=5.0in]{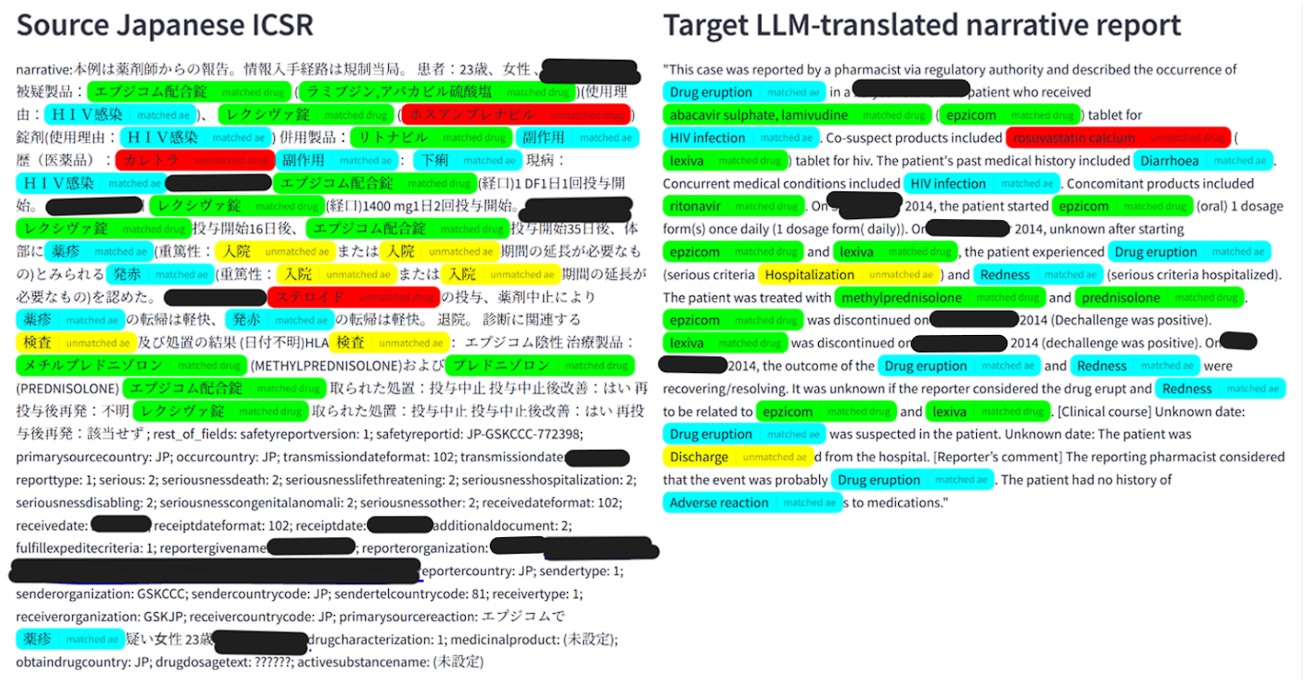}
			\caption{Illustration of guardrails filtering matched and unmatched drug terms and adverse event (AE) terms in the original Japanese ICSR and the LLM produced English case report. Text spans in blue indicate AEs that were successfully matched between the two texts while spans in yellow indicate AEs that were unmatched. Spans in green represent matched drugs while spans in red represent unmatched drugs. Sensitive information has been redacted with black bars.}\label{fig_03}
		\end{figure}

\begin{figure}
			\centering
			\includegraphics[width=5.0in]{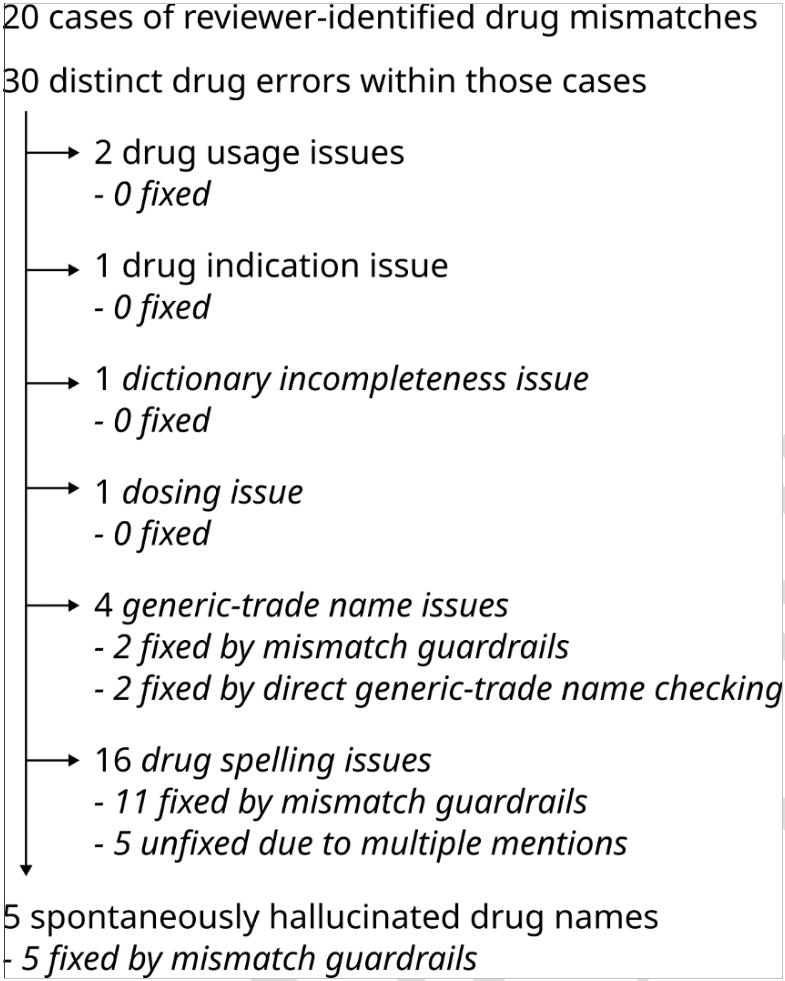}
			\caption{Counts of reviewer-identified drug error categories and mismatch guardrail fixes thereof. For each category, counts are given indicating which of the errors had been flagged.}\label{fig_04}
		\end{figure}

			Figure \ref{fig_03} shows an interface that illustrates the drug and AE MISMATCH guardrail. With this interface, unmatched entities are quickly highlighted, allowing downstream users of this system to understand the specific mismatches that would lead the system to re-route the case to automatic or human-in-the-loop adjudication, and for qualified users to identify and resolve specific issues. For a quantitative evaluation, we report the success rate of the MISMATCH guardrail in identifying one ``never event,'' a subset of human evaluator-identified drug issues, in a randomly selected set of 20 cases from the 210 that were evaluated. As can be seen in Figure \ref{fig_04}, all instances of the never event, ``spontaneously hallucinated drug names,'' were correctly identified by the MISMATCH guardrail.

			The missrates for the MISMATCH guardrail are summarized in Figure \ref{fig:s02} (comparing the model’s outputted target text to the source text) and Table \ref{tab:s2} (comparing the original standard’s target text to the source text). There were significant amounts of cases with a high ratio of unmatched adverse events, despite using the original standard source translations. Notwithstanding the reviewers' noted imperfection of those translations (see the Phase 1 section), the difference could be explainable by the increased number of ordinary words that are found in AEs. 
	
	\section{Discussion}

		Our investigation represents a significant step in the application of LLMs within PV, a field where accuracy and safety are paramount. We have explored one of the first integrations of LLMs into the PV workflow, particularly focusing on translating Japanese ICSRs to English. Through the deployment and critical assessment of both hard and soft semantic guardrails, our work confronts the critical challenges associated with LLMs, namely the propensity for hallucinations and the inherent uncertainties associated with model predictions. These approaches are complementary and therefore should be used in conjunction with other strategies to improve the quality of LLM outputs (e.g., temperature adjustments and prompt engineering). Even with safety-critical applications, there is variability in tolerance to inaccurate outputs: the impact of some issues could be so significant that safeguards are needed. In the context of LLM usage, safeguards could be guardrails in addition to or even before full human review. Our findings reveal that strategic guardrail applications effectively mitigate the risk of ``never event'' errors, with our MISMATCH guardrail successfully identifying every instance of hallucinated drug names in our translated texts from a carefully chosen case sample. We anticipate in routine usage as part of quality systems the ability to articulate a priori that certain errors cannot occur. We also note that some erroneous hallucinations could be so problematic that even if human review corrected them, the risk of wrongly recalling them as true outputs could still be problematic: the ability to remove such errors prior to human review holds advantages.

		Furthermore, we introduced both document-level and token-level uncertainty guardrails to facilitate a process that incorporates human oversight. The document-level guardrail serves to screen out irrelevant text, reducing unnecessary LLM processing at the ICSR intake stage, whereas the token-level guardrail flags segments of the generated text that exhibit low confidence. These measures immediately make outputs look less definitive and enable the rigorous verification of LLM outputs by skilled human evaluators, who can further investigate and rectify potential inaccuracies. Specifically, the token-level guardrail is designed to highlight areas of high entropy -- signifying considerable uncertainty -- for thorough review, thereby addressing potential inaccuracies extending beyond specific entities such as drug names or AEs. This approach adds to the burgeoning methodologies aimed at quantifying and communicating model uncertainties to users, supporting human-in-the-loop review and mitigation of risks.

		To our knowledge, this project is the first of its kind to develop and implement a range of guardrails for an LLM within the medical safety-sensitive environment of PV. As we look forward, we envision LLMs playing an increasingly central role in this sector, with ongoing improvements enhancing their precision and reliability. Nonetheless, the concept of never events, and its potential extrapolations into other medical safety critical areas, underscores a continuous need for robust guardrails like those we have developed here. The combination of LLMs with these safeguards offers a foundational model for their responsible and efficacious application in PV and beyond.

	\section{Limitations}

		Our work also has several limitations. We focused initial evaluations of hard guardrails on the problem of drug name hallucinations, but there are other kinds of errors that are classified as never events, like misinterpreting exposure outcomes of dechallenge/ rechallenge and AEs. Furthermore, although we did not solve for drug misspellings, this represents a type of error that may be addressed on case intake prospectively, while it could also be resolved by using structured data elements, retrospectively. Further work will extend the list of PV never events and their encoding in the system. Lastly, token-level uncertainty guardrails represent an area of evolving research and will likely continue to improve as the research field produces more solutions to quantify and informatively convey LLM output uncertainty.

	\section{Declarations}

		\textbf{Funding} GlaxoSmithKline Biologicals SA covered all costs associated with the 
			conduct of the study and the development of the manuscript and the decision to publish the manuscript. \\
	
		\textbf{Competing interests} All GSK co-authors receive GSK salary and some hold GSK stock 
			and stock options. Andrew L Beam is a consultant for Generate Biomedicines and Flagship Pioneering, Inc and holds stock and stock options in Generate Biomedicines and FL 85, Inc. \\
		
		\textbf{Author Contributions} JBH, JLP, and A. Beam contributed to the study concept, data
			 acquisition, data analysis, and data interpretation. DR, VK, and A. Bate contributed to the study concept, and data interpretation. GP contributed to data interpretation, and PS and CS contributed to the data analysis and data interpretation. \\
		
		\textbf{Data availability} The datasets generated and analyzed during the current study are not 	
			publicly available.\\		
		
		\textbf{Conflict of interests} This manuscript has not been submitted to, nor is
		under review at, another journal or other publishing venue. \\
	
		\section{Acknowledgments}
			
			The authors would like to thank the GSK staff from Japan safety and Global Safety who participated in phase 1 and 2 of the study; Phase 1: Asako Takata, Kohei Ogawa, Toshifumi Kimura, Tomoko Matsukawa, Kaoru Fujikura, Yoko Hijioka, Hiroko Toyota, Tamami Kaneko, Hiroki Nagahama, Takako Watanabe, Naoki Kaneko, Akiko Suhara; Phase 2: Tony Ning, Weronika Dardzinska, Marta Krzywdzinska-Kopacz, Joanna Kawałek, Chetan Sharma, Manju Uttam, Avinash Chaturvedula, Anupama Mathew, Aparna Jayachandra, Abhinaya Surender, Kaverappa M D, and Ewa Nowicka. In addition, students from the University of North Carolina, USA who supported Phase 2: Nathan Andert, Herbert Wan and Jackie Tan and GSK Global Safety Staff François Haguinet and Olivia Mahaux who supported with the multilingual assessments.

	\bibliography{harvard}


\begin{thebibliography}{28}
\ifx \bisbn   \undefined \def \bisbn  #1{ISBN #1}\fi
\ifx \binits  \undefined \def \binits#1{#1}\fi
\ifx \bauthor  \undefined \def \bauthor#1{#1}\fi
\ifx \batitle  \undefined \def \batitle#1{#1}\fi
\ifx \bjtitle  \undefined \def \bjtitle#1{#1}\fi
\ifx \bvolume  \undefined \def \bvolume#1{\textbf{#1}}\fi
\ifx \byear  \undefined \def \byear#1{#1}\fi
\ifx \bissue  \undefined \def \bissue#1{#1}\fi
\ifx \bfpage  \undefined \def \bfpage#1{#1}\fi
\ifx \blpage  \undefined \def \blpage #1{#1}\fi
\ifx \burl  \undefined \def \burl#1{\textsf{#1}}\fi
\ifx \doiurl  \undefined \def \doiurl#1{\url{https://doi.org/#1}}\fi
\ifx \betal  \undefined \def \betal{\textit{et al.}}\fi
\ifx \binstitute  \undefined \def \binstitute#1{#1}\fi
\ifx \binstitutionaled  \undefined \def \binstitutionaled#1{#1}\fi
\ifx \bctitle  \undefined \def \bctitle#1{#1}\fi
\ifx \beditor  \undefined \def \beditor#1{#1}\fi
\ifx \bpublisher  \undefined \def \bpublisher#1{#1}\fi
\ifx \bbtitle  \undefined \def \bbtitle#1{#1}\fi
\ifx \bedition  \undefined \def \bedition#1{#1}\fi
\ifx \bseriesno  \undefined \def \bseriesno#1{#1}\fi
\ifx \blocation  \undefined \def \blocation#1{#1}\fi
\ifx \bsertitle  \undefined \def \bsertitle#1{#1}\fi
\ifx \bsnm \undefined \def \bsnm#1{#1}\fi
\ifx \bsuffix \undefined \def \bsuffix#1{#1}\fi
\ifx \bparticle \undefined \def \bparticle#1{#1}\fi
\ifx \barticle \undefined \def \barticle#1{#1}\fi
\bibcommenthead
\ifx \bconfdate \undefined \def \bconfdate #1{#1}\fi
\ifx \botherref \undefined \def \botherref #1{#1}\fi
\ifx \url \undefined \def \url#1{\textsf{#1}}\fi
\ifx \bchapter \undefined \def \bchapter#1{#1}\fi
\ifx \bbook \undefined \def \bbook#1{#1}\fi
\ifx \bcomment \undefined \def \bcomment#1{#1}\fi
\ifx \oauthor \undefined \def \oauthor#1{#1}\fi
\ifx \citeauthoryear \undefined \def \citeauthoryear#1{#1}\fi
\ifx \endbibitem  \undefined \def \endbibitem {}\fi
\ifx \bconflocation  \undefined \def \bconflocation#1{#1}\fi
\ifx \arxivurl  \undefined \def \arxivurl#1{\textsf{#1}}\fi
\csname PreBibitemsHook\endcsname

\bibitem[\protect\citeauthoryear{Tang et~al.}{2023}]{tang2023evaluating}
\begin{barticle}
\bauthor{\bsnm{Tang}, \binits{L.}},
\bauthor{\bsnm{Sun}, \binits{Z.}},
\bauthor{\bsnm{Idnay}, \binits{B.}},
\bauthor{\bsnm{Nestor}, \binits{J.G.}},
\bauthor{\bsnm{Soroush}, \binits{A.}},
\bauthor{\bsnm{Elias}, \binits{P.A.}},
\bauthor{\bsnm{Xu}, \binits{Z.}},
\bauthor{\bsnm{Ding}, \binits{Y.}},
\bauthor{\bsnm{Durrett}, \binits{G.}},
\bauthor{\bsnm{Rousseau}, \binits{J.F.}}, \betal:
\batitle{Evaluating large language models on medical evidence summarization}.
\bjtitle{npj Digital Medicine}
\bvolume{6}(\bissue{1}),
\bfpage{158}
(\byear{2023})
\end{barticle}
\endbibitem

\bibitem[\protect\citeauthoryear{Singhal et~al.}{2022}]{singhal2022large}
\begin{botherref}
\oauthor{\bsnm{Singhal}, \binits{K.}},
\oauthor{\bsnm{Azizi}, \binits{S.}},
\oauthor{\bsnm{Tu}, \binits{T.}},
\oauthor{\bsnm{Mahdavi}, \binits{S.S.}},
\oauthor{\bsnm{Wei}, \binits{J.}},
\oauthor{\bsnm{Chung}, \binits{H.W.}},
\oauthor{\bsnm{Scales}, \binits{N.}},
\oauthor{\bsnm{Tanwani}, \binits{A.}},
\oauthor{\bsnm{Cole-Lewis}, \binits{H.}},
\oauthor{\bsnm{Pfohl}, \binits{S.}}, et al.:
Large language models encode clinical knowledge.
arXiv preprint arXiv:2212.13138
(2022)
\end{botherref}
\endbibitem

\bibitem[\protect\citeauthoryear{Clusmann et~al.}{2023}]{clusmann2023future}
\begin{barticle}
\bauthor{\bsnm{Clusmann}, \binits{J.}},
\bauthor{\bsnm{Kolbinger}, \binits{F.R.}},
\bauthor{\bsnm{Muti}, \binits{H.S.}},
\bauthor{\bsnm{Carrero}, \binits{Z.I.}},
\bauthor{\bsnm{Eckardt}, \binits{J.-N.}},
\bauthor{\bsnm{Laleh}, \binits{N.G.}},
\bauthor{\bsnm{L{\"o}ffler}, \binits{C.M.L.}},
\bauthor{\bsnm{Schwarzkopf}, \binits{S.-C.}},
\bauthor{\bsnm{Unger}, \binits{M.}},
\bauthor{\bsnm{Veldhuizen}, \binits{G.P.}}, \betal:
\batitle{The future landscape of large language models in medicine}.
\bjtitle{Communications medicine}
\bvolume{3}(\bissue{1}),
\bfpage{141}
(\byear{2023})
\end{barticle}
\endbibitem

\bibitem[\protect\citeauthoryear{Zhang et~al.}{2023}]{zhang2023siren}
\begin{botherref}
\oauthor{\bsnm{Zhang}, \binits{Y.}},
\oauthor{\bsnm{Li}, \binits{Y.}},
\oauthor{\bsnm{Cui}, \binits{L.}},
\oauthor{\bsnm{Cai}, \binits{D.}},
\oauthor{\bsnm{Liu}, \binits{L.}},
\oauthor{\bsnm{Fu}, \binits{T.}},
\oauthor{\bsnm{Huang}, \binits{X.}},
\oauthor{\bsnm{Zhao}, \binits{E.}},
\oauthor{\bsnm{Zhang}, \binits{Y.}},
\oauthor{\bsnm{Chen}, \binits{Y.}}, et al.:
{Siren's song in the AI ocean: a survey on hallucination in large language
  models}.
arXiv preprint arXiv:2309.01219
(2023)
\end{botherref}
\endbibitem

\bibitem[\protect\citeauthoryear{McKenna et~al.}{2023}]{mckenna2023sources}
\begin{botherref}
\oauthor{\bsnm{McKenna}, \binits{N.}},
\oauthor{\bsnm{Li}, \binits{T.}},
\oauthor{\bsnm{Cheng}, \binits{L.}},
\oauthor{\bsnm{Hosseini}, \binits{M.J.}},
\oauthor{\bsnm{Johnson}, \binits{M.}},
\oauthor{\bsnm{Steedman}, \binits{M.}}:
Sources of hallucination by large language models on inference tasks.
arXiv preprint arXiv:2305.14552
(2023)
\end{botherref}
\endbibitem

\bibitem[\protect\citeauthoryear{Xiong et~al.}{2023}]{xiong2023can}
\begin{botherref}
\oauthor{\bsnm{Xiong}, \binits{M.}},
\oauthor{\bsnm{Hu}, \binits{Z.}},
\oauthor{\bsnm{Lu}, \binits{X.}},
\oauthor{\bsnm{Li}, \binits{Y.}},
\oauthor{\bsnm{Fu}, \binits{J.}},
\oauthor{\bsnm{He}, \binits{J.}},
\oauthor{\bsnm{Hooi}, \binits{B.}}:
{Can LLMs express their uncertainty? An empirical evaluation of confidence
  elicitation in LLMs}.
arXiv preprint arXiv:2306.13063
(2023)
\end{botherref}
\endbibitem

\bibitem[\protect\citeauthoryear{Wagle et~al.}{2023}]{wagle2023empirical}
\begin{botherref}
\oauthor{\bsnm{Wagle}, \binits{S.}},
\oauthor{\bsnm{Munikoti}, \binits{S.}},
\oauthor{\bsnm{Acharya}, \binits{A.}},
\oauthor{\bsnm{Smith}, \binits{S.}},
\oauthor{\bsnm{Horawalavithana}, \binits{S.}}:
Empirical evaluation of uncertainty quantification in retrieval-augmented
  language models for science.
arXiv preprint arXiv:2311.09358
(2023)
\end{botherref}
\endbibitem

\bibitem[\protect\citeauthoryear{Bolton et~al.}{2024}]{bolton2024rambla}
\begin{botherref}
\oauthor{\bsnm{Bolton}, \binits{W.J.}},
\oauthor{\bsnm{Poyiadzi}, \binits{R.}},
\oauthor{\bsnm{Morrell}, \binits{E.R.}},
\oauthor{\bsnm{Bueno}, \binits{G.v.B.G.}},
\oauthor{\bsnm{Goetz}, \binits{L.}}:
{RAmBLA: A Framework for Evaluating the Reliability of LLMs as Assistants in
  the Biomedical Domain}.
arXiv preprint arXiv:2403.14578
(2024)
\end{botherref}
\endbibitem

\bibitem[\protect\citeauthoryear{{European Medicines Agency}}{2018}]{ema2018}
\begin{botherref}
\oauthor{\bsnm{{European Medicines Agency}}}:
{ICH E2B (R3) Electronic transmission of individual case safety reports
  (ICSRs): data elements and message specification implementation guide,
  Scientific guideline}
(2018)
\end{botherref}
\endbibitem

\bibitem[\protect\citeauthoryear{Bowen and Stavridou}{1993}]{bowen1993safety}
\begin{barticle}
\bauthor{\bsnm{Bowen}, \binits{J.}},
\bauthor{\bsnm{Stavridou}, \binits{V.}}:
\batitle{Safety-critical systems, formal methods and standards}.
\bjtitle{Software engineering journal}
\bvolume{8}(\bissue{4}),
\bfpage{189}--\blpage{209}
(\byear{1993})
\end{barticle}
\endbibitem

\bibitem[\protect\citeauthoryear{{World Health
  Organization}}{2002}]{world2002importance}
\begin{botherref}
\oauthor{\bsnm{{World Health Organization}}}:
The importance of pharmacovigilance
(2002)
\end{botherref}
\endbibitem

\bibitem[\protect\citeauthoryear{Bate and Evans}{2009}]{bate2009quantitative}
\begin{barticle}
\bauthor{\bsnm{Bate}, \binits{A.}},
\bauthor{\bsnm{Evans}, \binits{S.}}:
\batitle{Quantitative signal detection using spontaneous {ADR} reporting}.
\bjtitle{Pharmacoepidemiology and drug safety}
\bvolume{18}(\bissue{6}),
\bfpage{427}--\blpage{436}
(\byear{2009})
\end{barticle}
\endbibitem

\bibitem[\protect\citeauthoryear{Dong et~al.}{2024}]{dong2024building}
\begin{botherref}
\oauthor{\bsnm{Dong}, \binits{Y.}},
\oauthor{\bsnm{Mu}, \binits{R.}},
\oauthor{\bsnm{Jin}, \binits{G.}},
\oauthor{\bsnm{Qi}, \binits{Y.}},
\oauthor{\bsnm{Hu}, \binits{J.}},
\oauthor{\bsnm{Zhao}, \binits{X.}},
\oauthor{\bsnm{Meng}, \binits{J.}},
\oauthor{\bsnm{Ruan}, \binits{W.}},
\oauthor{\bsnm{Huang}, \binits{X.}}:
{Building Guardrails for Large Language Models}.
arXiv preprint arXiv:2402.01822
(2024)
\end{botherref}
\endbibitem

\bibitem[\protect\citeauthoryear{{Health Service
  Journal}}{2009}]{hsjNeverEvents}
\begin{botherref}
\oauthor{\bsnm{{Health Service Journal}}}:
{Guidance on implementing the never events framework}
(2009).
\url{https://www.hsj.co.uk/home/guidance-on-implementing-the-never-events-framework/5000691.article}
\end{botherref}
\endbibitem

\bibitem[\protect\citeauthoryear{Anderson and Watt}{2020}]{anderson2020using}
\begin{barticle}
\bauthor{\bsnm{Anderson}, \binits{J.E.}},
\bauthor{\bsnm{Watt}, \binits{A.J.}}:
\batitle{{Using Safety-II and resilient healthcare principles to learn from
  Never Events}}.
\bjtitle{International Journal for Quality in Health Care}
\bvolume{32}(\bissue{3}),
\bfpage{196}--\blpage{203}
(\byear{2020})
\end{barticle}
\endbibitem

\bibitem[\protect\citeauthoryear{{NQF}}{2024}]{nqfSres}
\begin{botherref}
\oauthor{\bsnm{{NQF}}}:
{List of SREs}
(2024).
\url{https://www.qualityforum.org/Topics/SREs/List_of_SREs.aspx}
\end{botherref}
\endbibitem

\bibitem[\protect\citeauthoryear{{European Medicines Agency}}{2017}]{ema2017}
\begin{botherref}
\oauthor{\bsnm{{European Medicines Agency}}}:
{Guideline on Good Pharmacovigilance Practices (GVP): Module VI – Collection,
  Management and Submission of Reports of Suspected Adverse Reactions to
  Medicinal Products (Rev 2). EMA/873138/2011 Rev 2}
(2017).
\url{https://www.ema.europa.eu/en/documents/regulatory-procedural-guideline/guideline-good-pharmacovigilance-practices-gvp-module-vi-collection-management-and-submission-reports-suspected-adverse-reactions-medicinal-products-rev-2_en.pdf}
\end{botherref}
\endbibitem

\bibitem[\protect\citeauthoryear{{FDA}}{2022}]{fdae2b}
\begin{botherref}
\oauthor{\bsnm{{FDA}}}:
{E2B(R3) Electronic Transmission of Individual Case Safety Reports
  Implementation Guide -- Data Elements and Message Specification; and Appendix
  to the Implementation Guide — Backwards and Forwards Compatibility}
(2022).
\url{https://www.fda.gov/regulatory-information/search-fda-guidance-documents/e2br3-electronic-transmission-individual-case-safety-reports-implementation-guide-data-elements-and}
\end{botherref}
\endbibitem

\bibitem[\protect\citeauthoryear{Zhang et~al.}{2020}]{zhang2020improving}
\begin{botherref}
\oauthor{\bsnm{Zhang}, \binits{B.}},
\oauthor{\bsnm{Williams}, \binits{P.}},
\oauthor{\bsnm{Titov}, \binits{I.}},
\oauthor{\bsnm{Sennrich}, \binits{R.}}:
Improving massively multilingual neural machine translation and zero-shot
  translation.
arXiv preprint arXiv:2004.11867
(2020)
\end{botherref}
\endbibitem

\bibitem[\protect\citeauthoryear{Graves}{2012}]{graves2012sequence}
\begin{botherref}
\oauthor{\bsnm{Graves}, \binits{A.}}:
Sequence transduction with recurrent neural networks.
arXiv preprint arXiv:1211.3711
(2012)
\end{botherref}
\endbibitem

\bibitem[\protect\citeauthoryear{Su et~al.}{2022}]{su2022contrastive}
\begin{barticle}
\bauthor{\bsnm{Su}, \binits{Y.}},
\bauthor{\bsnm{Lan}, \binits{T.}},
\bauthor{\bsnm{Wang}, \binits{Y.}},
\bauthor{\bsnm{Yogatama}, \binits{D.}},
\bauthor{\bsnm{Kong}, \binits{L.}},
\bauthor{\bsnm{Collier}, \binits{N.}}:
\batitle{A contrastive framework for neural text generation}.
\bjtitle{Advances in Neural Information Processing Systems}
\bvolume{35},
\bfpage{21548}--\blpage{21561}
(\byear{2022})
\end{barticle}
\endbibitem

\bibitem[\protect\citeauthoryear{Papineni et~al.}{2002}]{papineni2002bleu}
\begin{bchapter}
\bauthor{\bsnm{Papineni}, \binits{K.}},
\bauthor{\bsnm{Roukos}, \binits{S.}},
\bauthor{\bsnm{Ward}, \binits{T.}},
\bauthor{\bsnm{Zhu}, \binits{W.-J.}}:
\bctitle{{BLEU}: a method for automatic evaluation of machine translation}.
In: \bbtitle{Proceedings of the 40th Annual Meeting of the Association for
  Computational Linguistics},
pp. \bfpage{311}--\blpage{318}
(\byear{2002})
\end{bchapter}
\endbibitem

\bibitem[\protect\citeauthoryear{Post}{2018}]{post2018call}
\begin{botherref}
\oauthor{\bsnm{Post}, \binits{M.}}:
A call for clarity in reporting {BLEU} scores.
arXiv preprint arXiv:1804.08771
(2018)
\end{botherref}
\endbibitem

\bibitem[\protect\citeauthoryear{Klakow and Peters}{2002}]{klakow2002testing}
\begin{barticle}
\bauthor{\bsnm{Klakow}, \binits{D.}},
\bauthor{\bsnm{Peters}, \binits{J.}}:
\batitle{Testing the correlation of word error rate and perplexity}.
\bjtitle{Speech Communication}
\bvolume{38}(\bissue{1-2}),
\bfpage{19}--\blpage{28}
(\byear{2002})
\end{barticle}
\endbibitem

\bibitem[\protect\citeauthoryear{Kara et~al.}{2023}]{kara2023finding}
\begin{barticle}
\bauthor{\bsnm{Kara}, \binits{V.}},
\bauthor{\bsnm{Powell}, \binits{G.}},
\bauthor{\bsnm{Mahaux}, \binits{O.}},
\bauthor{\bsnm{Jayachandra}, \binits{A.}},
\bauthor{\bsnm{Nyako}, \binits{N.}},
\bauthor{\bsnm{Golds}, \binits{C.}},
\bauthor{\bsnm{Bate}, \binits{A.}}:
\batitle{{Finding Needles in the Haystack: Clinical Utility Score for
  Prioritisation (CUSP), an Automated Approach for Identifying Spontaneous
  Reports with the Highest Clinical Utility}}.
\bjtitle{Drug Safety}
\bvolume{46}(\bissue{9}),
\bfpage{847}--\blpage{855}
(\byear{2023})
\end{barticle}
\endbibitem

\bibitem[\protect\citeauthoryear{Likert}{1932}]{likert1932technique}
\begin{botherref}
\oauthor{\bsnm{Likert}, \binits{R.}}:
A technique for the measurement of attitudes.
Archives of psychology
(1932)
\end{botherref}
\endbibitem

\bibitem[\protect\citeauthoryear{Brown et~al.}{1999}]{brown1999medical}
\begin{barticle}
\bauthor{\bsnm{Brown}, \binits{E.G.}},
\bauthor{\bsnm{Wood}, \binits{L.}},
\bauthor{\bsnm{Wood}, \binits{S.}}:
\batitle{{The medical dictionary for regulatory activities (MedDRA)}}.
\bjtitle{Drug safety}
\bvolume{20}(\bissue{2}),
\bfpage{109}--\blpage{117}
(\byear{1999})
\end{barticle}
\endbibitem

\bibitem[\protect\citeauthoryear{{Google}}{2024}]{googleEvalModels}
\begin{botherref}
\oauthor{\bsnm{{Google}}}:
{Evaluating models | AutoML Translation Documentation}
(2024).
\url{https://cloud.google.com/translate/automl/docs/evaluate}
\end{botherref}
\endbibitem

\end{thebibliography}

	\clearpage
	
	\FloatBarrier
	\section*{Tables and Figures}

	\FloatBarrier	
	\begin{table}[h]
		\caption{Numbers of individual case safety reports (ICSRs) and direct translation pairs}
		\label{t1}
		\begin{tabular}{ll}
			\toprule
			& \textbf{Number of ICSRs}  \\
			\midrule
			Pretraining examples (GSK), total & 131,037 \\
			ICSRs in Japanese & 	58,855\\
			ICSRs in Spanish & 13,264\\
			ICSRs in German & 30,370\\
			ICSRs in French & 28,548\\
			Japanese direct translation pairs (OPUS-100) & 10,000\\
			Spanish direct translation pairs (OPUS-100) & 10,000 \\
			German direct translation pairs (OPUS-100) & 10,000\\
			French direct translation pairs (OPUS-100) & 10,000\\
			\botrule
		\end{tabular}
	\end{table}
	
	\begin{table}[h]
		\caption{Per-token perplexity scores on held out data in the validation set, before fine-tuning (base model) and after fine-tuning on a parallel language corpus (fine-tuned models)}
		\label{t2}
		\begin{tabular}{ll}
			\toprule
			& \textbf{Perplexity}  \\
			\midrule
			\textbf{Base model} & \\
	
				mt5-xl & 2.72 × 103 \\
				mpt-7B instruct & 2.20 × 107 \\
				stablelm-japanese & 1.09 × 106  \\
			\textbf{Fine-tuned models} & \\ 
				mt5-xl  & 1.43 \\
				mpt-7B instruct & 113 \\
				stablelm-japanese & 131 \\
			\botrule
		\end{tabular}
	\end{table}

	\renewcommand{\arraystretch}{1.5}
	\begin{table}[h]
		\caption{Phase 2 frequencies of each error type in large language model (LLM) generated target text, as determined by human drug safety experts. A score of 5 in each category means the target text was essentially without error, 4 indicates minor errors that do not affect interpretation, 3 indicates errors that might have a small impact on interpretation, 2 indicates an error that would change interpretation, and 1 indicates an unacceptable error. See Supplementary Table \ref{tab:s1} for a mapping of the score to the specific questions presented to the human reviewers.}
		\label{t3}
		\begin{tabular}{@{}p{5cm}ccccc@{}}
		\toprule
			& \multicolumn{5}{c}{\textbf{Score}}  \\	
			\midrule
	
			\textbf{Evaluation criteria} & \textbf{5} & \textbf{4}& \textbf{3}& \textbf{2} & \textbf{1} \\
	
				Is the original translation provided by the human clear? & 
				119 (56.7\%) & 
				72 (34.3\%) & 
				16 (7.6\%) & 
				3 (1.4\%) & 
				0 (0.0\%)  \\
				Is LLM translation clear? &
					32 (15.0\%) & 
					98 (46.7\%) & 
					56 (26.7\%) & 
					21 (10.0\%) & 
					3 (1.4\%) \\
				Is the LLM translation complete? &
					82 (39.0\%) & 
					70 (33.3\%) & 
					37 (17.6\%) & 
					21 (10.0\%) & 
					0 (0.0\%) \\
				Is the information in the LLM translation correct? &
					19 (9.0\%) & 
					68 (32.4\%) & 
					91 (43.3\%) & 
					32 (15.2\%) & 
					0 (0.0\%) \\
				Is there unnecessary or extraneous information in the LLM translation? &
					97 (46.2\%) & 
					78 (37.1\%) & 
					28 (13.3\%) & 
					3 (1.4\%) & 
					4 (1.9\%) \\
				Amount of key* (drug safety related) information in the LLM translation not present in the source text &
					108 (51.4\%) & 
					48 (22.9\%) & 
					36 (17.1\%) & 
					11 (5.2\%) & 
					7 (3.3\%) \\
	
			\botrule
		\end{tabular}
	\end{table}

	\renewcommand{\arraystretch}{1}

	\renewcommand{\arraystretch}{1.5}
	\begin{table}[h]
		\caption{Phase 2 fine-grained assessment of the presence of any error in the target text for several important error categories.}
		\label{t4}
		\begin{tabular}{lr@{}}
			\toprule
			\textbf{Error Category} &   \textbf{Number (\%)} \\
			\midrule
			Source contains contradictions    & 30 (14\%) \\
			LLM contains contradictions   & 86 (41\%) \\
			Wrong drug name or information  & 127 (60\%) \\
			Wrong dosage   & 34 (16\%) \\
			Wrong dates/times  & 135 (64\%) \\
			Incorrect/missing AE/wrong outcome    & 149 (71\%) \\
			Rechallenge/dechallenge errors & 13 (6\%) \\
			TTO issues & 48 (23\%) \\
			Nonsensical phrases & 135 (64\%) \\
			Other errors & 157 (75\%) \\
			Is the case clinically accurate?    & 73 (35\%)  \\
			\midrule
			\multicolumn{2}{l}{\textit{Abbreviations: LLM = large language model; AE = adverse event; TTO = time to onset.}} \\
			\bottomrule
		\end{tabular}
	\end{table}

	\renewcommand{\arraystretch}{1}

	\FloatBarrier
	\section*{Supplement}
		
	\setcounter{table}{0}
	
	\renewcommand{\thetable}{S\arabic{table}}
		
	\FloatBarrier		
	\renewcommand{\arraystretch}{1.5}
	\begin{table}[h]
		\caption{Evaluators' key for five-point scales, used in human review of large language model (LLM) and original standard translations \cite{likert1932technique}.}

		\label{tab:s1}
		\begin{tabular}{@{}p{2.25cm}p{2.25cm}p{2.25cm}p{2.25cm}p{2.25cm}p{2.25cm}@{}}
			\toprule
			& \multicolumn{5}{c}{\textbf{Score}}  \\	
			\midrule
			
			 & \textbf{5} & \textbf{4}& \textbf{3}& \textbf{2} & \textbf{1} \\
	
				Is the original translation provided by the human clear? &
				Completely easily understood and well written &
				Mostly clear and easy to read &
				Needs rereading to understand &
				Difficult to understand &
				Unintelligible \\
			\midrule			
				Is LLM translation clear?  &
				Completely easily understood and well written  &
				Mostly clear and easy to read  &
				Needs rereading to understand  &
				Difficult to understand  &
				Unintelligible \\
			\midrule
				Is the LLM translation complete? &
				Complete, not missing any relevant or auxiliary information &
				Mostly clear and easy to read &
				Needs rereading to understand &
				Difficult to understand &
				Unintelligible \\
			\midrule			
				Is the information in the LLM translation correct? &
				All translated text correct   &
				Some incorrectness, no impact to interpretation   &
				Inaccuracy that might impact interpretability   &
				Significant inaccuracies impacting interpretability   &
				All translation is inaccurate   \\
			\midrule			
				Is there unnecessary or extraneous information in the LLM translation? &
				No extra information in LLM text &
				Little extra information, none impacting interpretation &
				Some extra information that might affect the case interpretation &
				Significant extra information but not all changes interpretation &
				Significant extra information changing interpretation \\
			\midrule			
				Amount of key* (drug safety related) information in the LLM translation not present in the source text &
				No extra information in LLM text &
				Little extra information, none impacting interpretation &
				Some extra information that might affect the case interpretation &
				Significant extra information but not all changes interpretation &
				Significant extra information changing interpretation \\
		
					\botrule
	\end{tabular}
	\end{table}

	\begin{table}[h]
		\caption{Summary statistics on the phase 1 assessments that aimed to evaluate whether the original standard version of the English target text sufficiently captured the same meaning as the original Japanese ICSR. A total of 210 cases underwent binary review, comparing content present in original standard source text to content present in Japanese ICSR.}
		\label{tab:s2}
		\begin{tabular}{ll}
			\toprule
			& \textbf{Number of cases (\%)}\\
			\midrule
				Correct assessment? & 125 (60\%) \\
				Source does not contain contradictions	& 209 (99\%) \\
			\botrule
		\end{tabular}
	\end{table}

	\begin{table}[h]
		\caption{Five-point scale review criteria of phase 1 assessment, comparing content present in original standard source text to content present in Japanese ICSR.}
		\label{tab:s3}
		\begin{tabular}{p{2.5cm}|p{1.5cm}|p{1.5cm}|p{1.5cm}|p{1.5cm}|p{1.5cm}}
			\toprule
	
				& Completely easily understood and well written &
				Mostly clear and easy to read &
				Needs rereading to understand &
				Difficult to understand &
				Unintelligible \\
			\midrule
				Source is clear: \mbox{primary reviewer} & 	77 & 102 & 26 &	5 & 0 \\
			\midrule			
				Source is clear: \mbox{secondary reviewer} & 70 & 118 & 20 & 2 & 0 \\
			\botrule
		\end{tabular}
	\end{table}

	\begin{table}[h]
		\caption{Phase 1 evalulation translation accuracy. Translation accuracy comparing content present in original standard provided source text to content present in target text, the Japanese ICSR. }
		\label{tab:s4}
		\begin{tabular}{p{1.5cm}|p{1.75cm}|p{1.75cm}|p{1.75cm}|p{1.75cm}}
			\toprule
			& Added information &
			Nothing missing or added to translation &
			Both with missing or added information & Missing information \\
			\midrule
			Translation accuracy & 140 & 29 & 29 & 12 \\
			\botrule
		\end{tabular}
	\end{table}

	\FloatBarrier

	\setcounter{figure}{0}
	\renewcommand{\thefigure}{S\arabic{figure}}
	
	\begin{figure}
		\centering
		\includegraphics[width=5.0in]{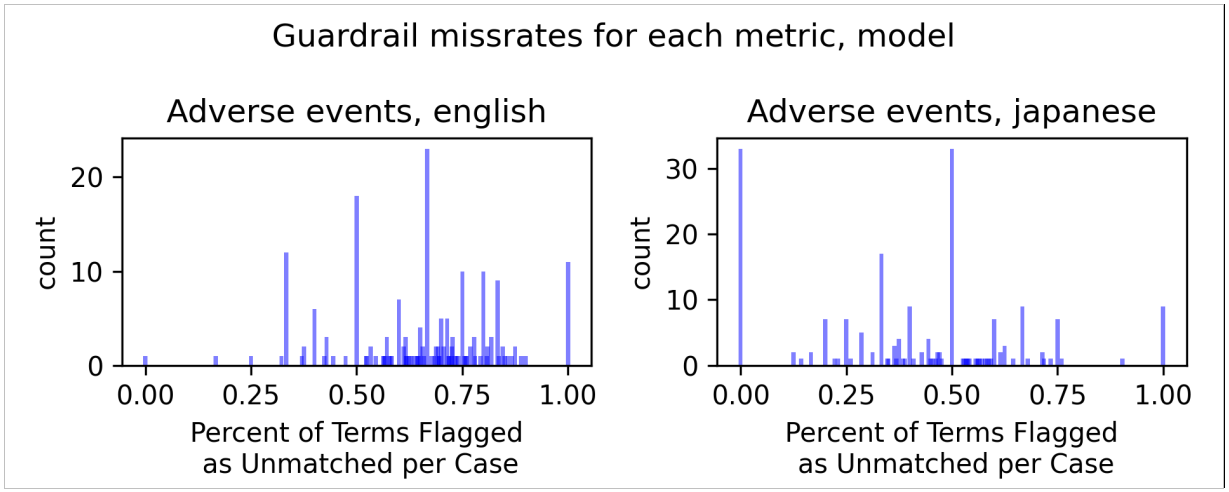}
		\caption{Guardrail missrates for adverse events in the original standard translation. Left: missrate comparing model output's target text and original source text, counting missing English adverse events (in the target but not the source). Right: comparing model output's target text and original source text, counting missing Japanese adverse events (in the source but not the target). }\label{fig:s01}
	\end{figure}
	
	\begin{figure}
		\centering
		\includegraphics[width=5.0in]{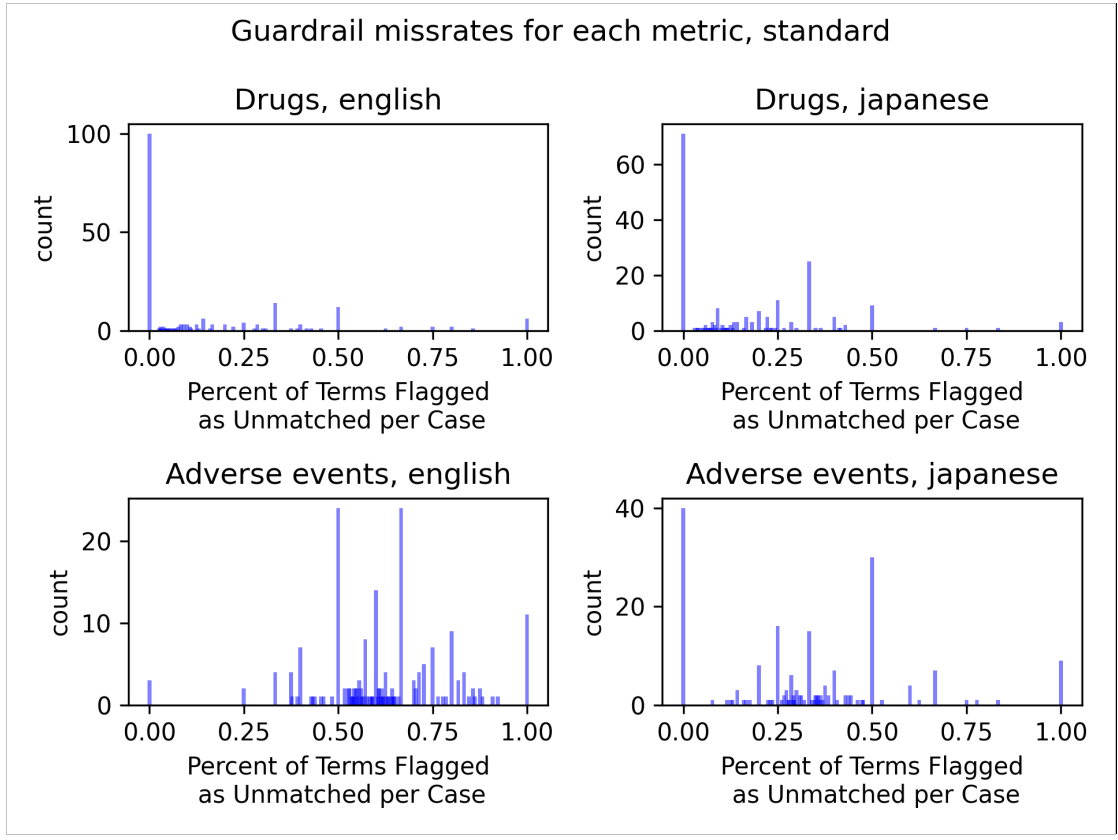}
		\caption{Histogram of missrates matching the original standard translation as a standard to the source Japanese text. Left: missrate comparing standard translation's target text and original source text, counting missing English drugs (top) and adverse events (bottom) (in the target but not the source). Right: comparing standard translation’s target text and original source text, counting missing Japanese drugs (top) and adverse events (bottom) (in the source but not the target).}\label{fig:s02}
	\end{figure}

\clearpage

	\section*{Preliminary feasibility study}
	
	In our pilot assessment of 20 translations, we found that 16/20 (80\%) were deemed as acceptable overall. Table \ref{tab:s5} shows a breakdown of the kinds of errors identified by human experts on this pilot dataset. The most common kinds of mistakes were miscellaneous errors, which includes misspellings and grammatical errors.  
	
	\begin{table}[h]
		\caption{Preliminary feasibility study, frequencies of each error type in LLM generated target text as determined by human drug safety expert}
		\label{tab:s5}
		\begin{tabular}{ll}
			\toprule
			\textbf{Error Type} & \textbf{Count (\%)} \\
			\midrule
				Incorrect drug name or information & 2 (10\%) \\
				Incorrect dosage (could be dose administered, dosing frequency, or dosage changes).  & 1 (5\%) \\
				Incorrect dates or times & 0 (0\%) \\
				Incorrect or missing adverse event or outcome & 3 (15\%) \\
				In correct rechallenge/dechallenge & 0 (0\%) \\
				Time to onset issues & 1 (5\%) \\
				Grammatical error & 5 (25\%) \\
				Miscellaneous & 9 (45\%) \\
				Unacceptable overall & 4 (20\%) \\
			
			\botrule
		\end{tabular}
	\end{table}

\end{document}